\documentclass[letterpaper,twocolumn,10pt,anonymous]{article}
\usepackage{PrivPRISM/style}

\usepackage{graphicx}
\usepackage[most]{tcolorbox}
\usepackage{listings}
\usepackage{xcolor}
\tcbuselibrary{listingsutf8}
\newtcblisting{promptboxlisting}[1][]{
  listing only,
  listing options={
    basicstyle=\ttfamily\small,
    breaklines=true,
    breakindent=0pt,
    showstringspaces=false,
    columns=fullflexible,
    escapeinside={(*@}{@*)}
  },
  colback=gray!10, 
  colframe=gray!80, 
  coltitle=white, 
  colbacktitle=gray!80, 
  title=#1,
  boxrule=0.5pt,
  arc=2pt,
  outer arc=2pt,
  top=1mm,
  bottom=1mm,
  left=2mm,
  right=2mm,
  before skip=10pt,
  after skip=0pt,
  enhanced,
  sharp corners=south,
  width=\linewidth
}

\begin{document}

\date{}

\title{\Large \bf PrivPRISM: Automatically Detecting Discrepancies Between Google Play Data Safety Declarations and Developer Privacy Policies}

\def\plainauthor{Author name(s) for PDF metadata. Don't forget to anonymize for submission!}

\author{
{\rm Bhanuka Silva}\\
\small bpin9254@sydney.edu.au\\
\small University of Sydney
\and
{\rm Dishanika Denipitiyage}\\
\small dden5444@sydney.edu.au\\
\small University of Sydney
\and
{\rm Anirban Mahanti}\\
\small anirban.mahanti@sydney.edu.au\\
\small University of Sydney
\and
{\rm Aruna Seneviratne}\\
\small a.seneviratne@unsw.edu.au\\
\small University of New South Wales
\and
{\rm Suranga Seneviratne}\\
\small suranga.seneviratne@sydney.edu.au\\
\small University of Sydney
} 

\maketitle

\begin{abstract}
End-users seldom read verbose privacy policies, leading app stores like Google Play to mandate simplified data safety declarations as a user-friendly alternative. However, these self-declared disclosures often contradict the full privacy policies, deceiving users about actual data practices and violating regulatory requirements for consistency. To address this, we introduce PrivPRISM, a robust framework that combines encoder and decoder language models to systematically extract and compare fine-grained data practices from privacy policies and to compare against data safety declarations, enabling scalable detection of non-compliance. 
Evaluating 7,770 popular mobile games uncovers discrepancies in nearly 53\% of cases, rising to 61\% among 1,711 widely used generic apps. Additionally, static code analysis reveals possible under-disclosures, with privacy policies disclosing just 66.8\% of potential accesses to sensitive data like location and financial information, versus only 36.4\% in data safety declarations of mobile games. Our findings expose systemic issues, including widespread reuse of generic privacy policies, vague / contradictory statements, and hidden risks in high-profile apps with 100M+ downloads, underscoring the urgent need for automated enforcement to protect platform integrity and for end-users to be vigilant about sensitive data they disclose via popular apps.
\end{abstract}

\section{Introduction}
\label{Sec:Introduction}

\begin{figure}[H]
    \centering
    \includegraphics[width=0.47\textwidth]{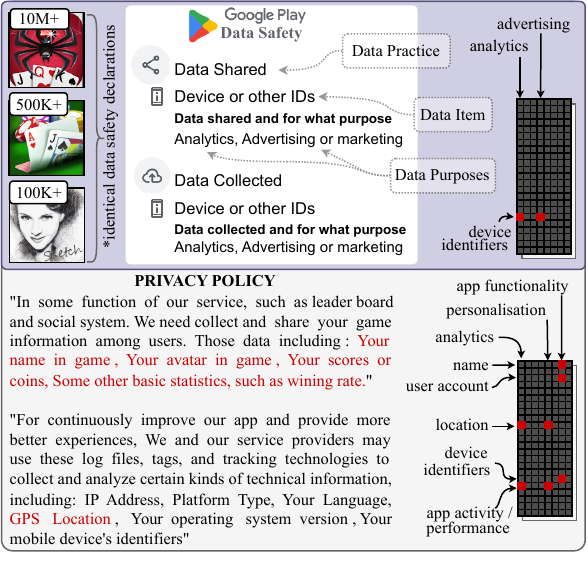}
    \caption{Top: Data safety declarations of three games by \emph{Spider Solitaire Card Games} and bottom: corresponding privacy policy highlighting inconsistencies. Data safety labels only report collecting and sharing device identifiers for analytics and ads, yet their privacy policy omit ad-related details and instead disclose access to more sensitive data such as GPS location, user accounts, and app activity. \emph{Identified via PrivPRISM framework deployment to mobile games.}}
    \label{fig:first_example}
\end{figure}

Mobile application (app) developers are now mandated to self-declare summarised labels of their data privacy practices, referred to as `data safety' (DS) on Android~\cite{Google_privacy_policy} or `app privacy’ on iOS~\cite{Apple_privacy_policy} (see Fig.~\ref{fig:first_example} for an illustration). These declarations offer a more user-friendly approach compared to privacy policies (PPs), which are often criticised for being lengthy and complicated~\cite{2008_cost_of_reading, www21_PP_overtime, why_ignore_pp}.

However, they are not intended to replace the role of full privacy policies. End-users are still required to refer them for detailed information about the data practices and relevant purposes expanding beyond the exhaustive list of summarised labels. Regulatory frameworks such as General Data Protection Regulation (GDPR) for EU, California Consumer Privacy Act (CCPA) for US and Australian Privacy Principles (APPs), along with platform operators, require compliance in consistency across both forms of disclosure. Despite this, a small-scale manual investigation by the Mozilla Foundation involving 40 apps revealed significant discrepancies between PP and DS declarations in nearly 40\% of cases~\cite{caltrider2023seenoevil}. 
Given the vast number of apps on platforms like the Google Play Store, it is important to assess the extent to which such inconsistencies persist at scale between PP and DS declarations, both self-declared by developers, and the implications they have for end-user privacy.

Regulatory enforcement against non-compliant apps is typically reactive, triggered by end-user complaints, which highlights the need for automated compliance verification. However, existing methods suffer from two major shortcomings. First, they do not evaluate privacy policies against data-safety declarations; both of which are developer self-attestations and lack metrics for comparing consistency across the two. Second, they depend entirely on either embedding-based or autoregressive language models for text analysis, limiting the frameworks with the disadvantages of each category. For example, it is well known that semantic search on embedding-based models is noisy \cite{gao2023retrieval} while the precision on decoder-based information retrieval is comparatively lower \cite{silva2024entailment}.

To address this, we introduce PrivPRISM  (\underline{Pri}vacy \underline{P}olicy \underline{R}easoning and \underline{I}nvestigation using \underline{S}ystematic language-\underline{M}odelling), a novel framework that leverages both encoder and decoder based language models for fine-grained extraction and verification of data practices from privacy policies. Our major contributions include;

\begin{itemize}
    \item \textbf{PrivPRISM framework for robust compliance checks:} Benchmarking demonstrates PrivPRISM’s superior performance, achieving a 6 percentage points higher precision than state-of-the-art GPT baselines in data practice classification and reducing mapping errors by 22.3 percentage points.
    \item \textbf{Large-scale empirical analysis\footnote{\href{https://github.com/NSS-USYD/PrivCORPUS}{Data repository - github.com/NSS-USYD/PrivCORPUS}}:} We apply PrivPRISM to 7,770 popular mobile games on the Google Play Store, comparing extracted PP data practices against DS declarations using tailored compliance metrics. Our findings, including high-profile cases with 100M+ downloads, reveal serious shortcomings in current self-disclosure practices.  Fig.~\ref{fig:first_example} portrays an example finding where three popular games (10M+ downloads) under-declare in their DS labels. We uncover that 53\% of apps exhibit such PP to DS inconsistencies (rising to 61\% in our evaluation of 1,711 non-game apps), with developers often providing vague purposes, inaccessible or mismatched policies, and contradictory statements, all of which undermine user trust.

    \item \textbf{Uncovering the disclosure gaps:} We observe that 64.9\% of apps reuse privacy policies, with 13.4\% PPs shared across 2-10 titles, obscuring app-specific data practices. Code-level analysis shows that while PPs account for only 66.8\% of potential sensitive data requests such as location, financial and user-account access, DS declarations cover merely 36.4\%, revealing a major compliance gap. A manual audit of 50 apps found that 38\% of policy URLs require redirection before reaching the actual policy page, violating Google Play policies and we discuss case studies highlighting ambiguous and contradicting policy texts that not only fail regulatory expectations but also expose end-users to hidden risks.

\end{itemize}

The organisation of this paper is as follows. Sec.~\ref{Sec:RelatedWork} discusses the literature and Sec.~\ref{Sec:Method} introduces the necessary terminology we use in the rest of the paper along with the PrivPRISM framework. Sec.~\ref{sec:metrics} highlights the tailored  metrics we use to quantitatively characterise the compliance landscape of mobile apps. Sec.~\ref{Sec:benchmarking} provides the benchmarking results for the PrivPRISM where we show the robustness and effectiveness of the framework followed up by the results in Sec.~\ref{sec:results}. First we analyse the results for 7,770 mobile games and next we generalise PrivPRISM framework by applying to 1,711 non-game (generic) apps. This section also elaborates an in-the-wild analysis with several selected case studies among popular developers. We conclude the paper in Sec.~\ref{sec:conclusion} and additionally redirect readers to Appendix for supplementary details.

\section{Related Work}
\label{Sec:RelatedWork}

Despite the increased concerns about online privacy, empirical studies often showcased that privacy policy documents are not given attention by end-users \cite{aussies_do_not_read, Auxier2019USstats, Black2018UKstats}. Regulatory actions to improve transparency has also inadvertently caused privacy policies to grow even large, diverging end-users from reading them even more \cite{take_some_cookies}. Existing research related to privacy policies highlight that even if end-users attempt to read them, they are far from perfect, consisting internal contradictions, non-disclosures and compliance gaps \cite{policylint, story2019natural, policy_landscape}.

\subsection{Privacy Labels in App Markets}

The introduction of iOS App Privacy labels in 2020, aimed to provide a user-friendlier disclosure label concept within the app metadata pages. Following this, new challenges emerged and survey-based studies by \cite{li2022understanding, zhang2022usable} and more recently \cite{keswani2025user} revealed developer misunderstandings and usability issues, while \cite{koch2022keeping} found violations of apps transmitting data without proper disclosure in app privacy labels. Android also launched Data Safety declaration labels for its apps in 2022. Khandelwal et al.\cite{khandelwal2023comparing} conducted a large-scale measurement of over 100k apps cross-listed on both platforms, uncovering a 60\% inconsistency rate where the same app reported different privacy practices on iOS versus Android.

More recently, scrutiny of Android's Data Safety section has revealed pervasive under-reporting. For instance, Arkalakis et al.~\cite{arkalakis2024abandon} conducted a longitudinal dynamic analysis of nearly 5,000 Android apps, finding that 81\% misrepresented their data practices in the Data Safety sections, with the majority (79.4\%) not asking the end user to provide consent for the data they collect and share. Complementing this, Khedkar et al.~\cite{khedkar2024android} utilised static analysis to show that developers frequently struggle to accurately report data collection due to Google's abstract definitions and a lack of automated tools to identify data collected by system APIs.

The primary limitation of the existing research on data safety declarations is that they do not provide a comprehensive framework to evaluate against the developer declared privacy policies, effectively making mobile app privacy policy and data safety based inconsistency evaluations to be treated as two separate branches, resulting in having no evaluation criteria to assess the cross-declaration consistency. In addressing this limitation, we detect disclosure-disclosure discrepancies while providing tailored metrics to quantify and clearly characterise them. We additionally compare against the potential evidence we can extract from app source codes to further elaborate on our findings. 

\subsection{NLP for Policy Understanding}

Traditional NLP techniques have long been applied to simplify privacy policies through summarisation \cite{gopinath2020automatic}, to support user decision-making \cite{sathyendra2016, sathyendra2017identifying}, answer user queries \cite{harkous2018}, and to build user-interactive privacy tools \cite{zimmeck2014privee, nokhbeh2020privacycheck}. The adoption of language models further advanced this space, with encoder-based architectures such as PrivBERT by \cite{privbert} and related transformer models by Adhikari et al. \cite{adhikari2023evolution} achieving state-of-the-art performance in tasks of data practice classification and question answering. PoliGraph~\cite{cui2023poligraph} is an example where RoBERTa encoder model has been adapted as a tool to capture statements in a privacy policy as relations between different parts of the text, for example, what is collected and who (what entity) receives this information.

More recently, generative models such as GPT and open-source counterparts have demonstrated competitive zero- and few-shot performance on privacy policy text classification tasks \cite{tang2023policygpt, rodriguez2024largelanguagemodelsnew, silva2024entailment, chanenson2025automating}.
LLM-powered tools, including CLEAR \cite{chen2025clear}, PRISMe \cite{freiberger2025you}, and Privacify \cite{woodring2024enhancing} demonstrate the potential of leveraging off-the-shelf generative models such as ChatGPT, GPT4o, Gemini and Mistral-7B to enhance end-user understanding of privacy policies and raise awareness of associated risks. 

Despite substantial progress in privacy-policy analysis, most existing tools that rely on latest advancements on language modelling are not designed for mobile-app–specific information extraction—an important gap given the dominance of mobile platforms in global digital activity; Android alone holding 43.1\% of the OS market share \cite{statcounter2025osshare}. As encoder and decoder models have unique limitations such as embedding space noise or low precision text generation \cite{gao2023retrieval, silva2024entailment}, prior work also do not examine how encoder and decoder models can be jointly leveraged to achieve fine-grained, semantically coherent interpretations of data practices. 

In contrast, our method uses a systematic encoder–decoder architecture to generate structured, semantically rich policy representations and detect inconsistencies with Google Play Data Safety declarations. It is integrated into a mobile-app–focused analysis framework that supports scalable, fully automated compliance assessment.

\section{PrivPRISM Framework}
\label{Sec:Method}

\begin{figure}[h]
    \centering
    \includegraphics[width=0.47\textwidth]{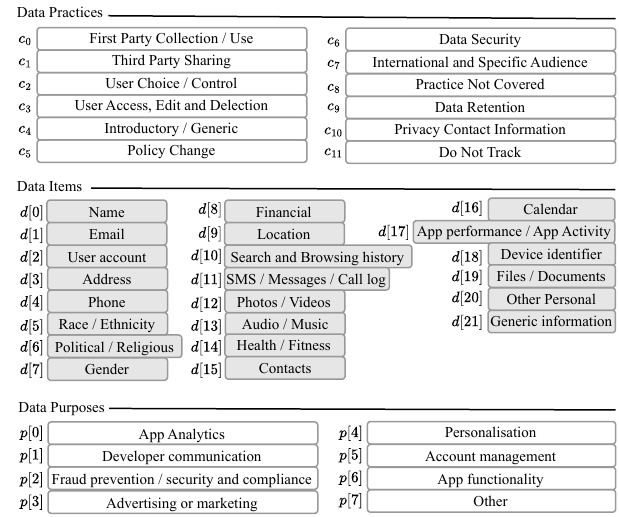}
    \caption{Terminology used in PrivPRISM}
    \label{fig:terminology}
\end{figure}

\begin{figure*}[t]
    \centering
    \includegraphics[width=0.97\textwidth]{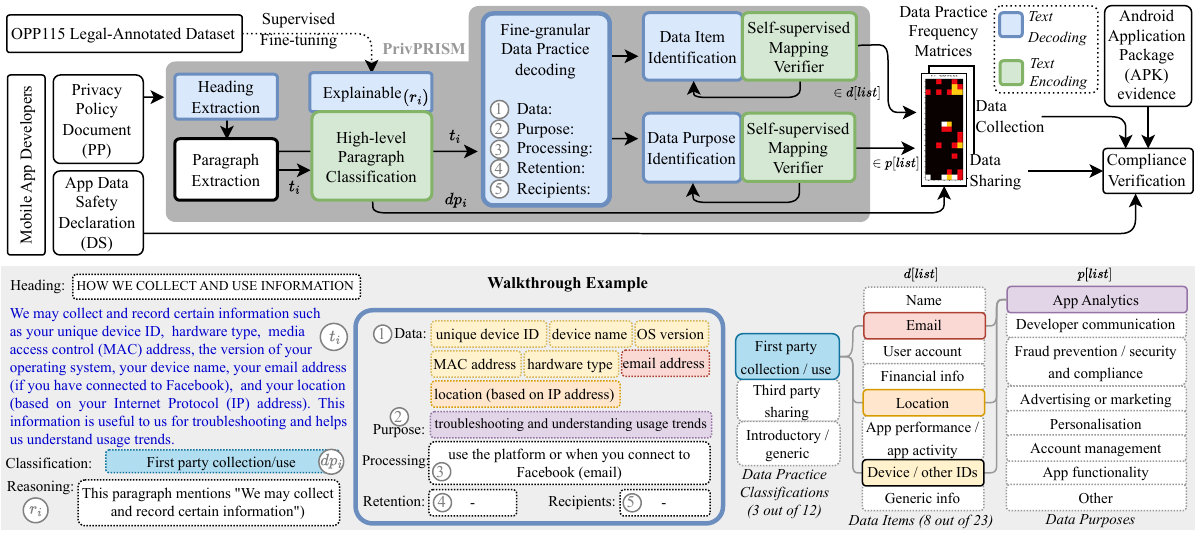}
    \caption{End-to-end pipeline of our framework (top) and a walkthrough example (bottom)}
    \label{fig:pipeline}
\end{figure*}

We adopt terminology consistent with Google’s definitions, a \textit{Data practice} is an activity involving data collection or sharing by a developer, a \textit{Data item} is a specific piece of information about the user (e.g., name, location, email) or the device (e.g., device ID, diagnostics) and a \textit{Data purpose} is the intended reason for processing a data item, such as analytics, or advertising. Fig.~\ref{fig:terminology} summarises the terminology we use throughout this paper and Figures~\ref{fig:first_example} and~\ref{fig:pipeline} illustrate how these definitions apply to data safety and privacy policy through examples. We redirect the readers to Appendix Sec.~\ref{subsec:appendix_DS} for further insights on the terminology we use.
PrivPRISM, as illustrated in Fig.~\ref{fig:pipeline}, is a systematic and verifiable privacy policy information extraction framework consisting of encoder and decoder models. 
We next describe various modules of PrivPRISM.

\subsection{Heading and Paragraph Extraction}

Section headings in a privacy policy, as defined by the developers, are essential structural cues, often summarising the content of the corresponding sections that follow. However, extracting these headings is non-trivial due to the lack of standardised HTML formatting across developers \cite{opt_outs_from_policy, harkous2018}.
To address this, we employ a generative language model (LlaMA3.1-8B-Instruct, 128k-token context window), prompting them with the textual body extracted from downloaded HTML file (see Appendix Sub.Sec.~\ref{subsec:appendix_dataset} for details) to identify and extract primary section headings.

Following heading extraction, we segment each policy into sections bounded by consecutive heading pairs. To ensure robust separation of main sections we run three independent heading extraction trials per policy and select the one with the highest mean and lowest standard deviation in section lengths. Sections may contain multiple paragraphs, separated by newline; short paragraphs (< 512 characters, $\sim$ 64 words) are merged with the following paragraphs. These merged paragraph texts ($t_i$ in Fig.~\ref{fig:pipeline}) often consolidate subheadings with their accompanying content, yielding more coherency.

\subsection{Explained Data Practice Classification}
\label{subsec:intro_exp_cls}

This module classifies a given paragraph $t_i$ into a data practice category, such as \textit{data collection, data sharing, other, etc.} OPP-115~\cite{opp115}, a popular dataset annotated by legal experts, is used as a benchmark for privacy policy paragraph classification~\cite{harkous2018} and using of fine-tuned encoder-based language models has already shown promising results \cite{privbert}. PolicyGPT \cite{tang2023policygpt} attempted this task using zero-shot GPT prompting. Recent work~\cite{silva2024entailment} emphasises that decoder model-based classification can introduce explainability to this task in alignment with expert annotations. Nonetheless, despite being user-friendly, this framework did not reach the accuracy levels of encoder-based models. Additionally, it could not filter the most confident class label output.

We propose a modified version of this idea, to first predict the most confident label $dp_i \in [c_0, c_1, .. c_{11}]$ for a paragraph $t_i$ via an encoder model (PrivBERT) and next to use a decoder model (Llama3.1-8B-Instruct) to provide explanations (text excerpt $r_i \in t_i$). The textual class label output of the encoder model acts as a prior for the decoder model's next word prediction objective, and we fine-tuned both models using the OPP-115 dataset. \textcolor{black}{This design, confirms that the encoder drives classification accuracy while the decoder adds interpretability.} Labels $c_o$ (first party collection) to $c_{11}$ (do-not-track) are consistent with prior work. Benchmark results for this module are summarised in Sub.Sec.~\ref{subsec:benchmark_exp_cls}.  

\subsection{Fine Granular Data Practice Decoding}

Based on the terminology provided by the minimum core model in \cite{bonatti2018special}, we extract five elements from a selected policy paragraph $t_i$; \textit{1. Data} or what is processed by a data practice operation, \textit{2. Purpose} of the operation, \textit{3. Processing} or the description of the operation (e.g., disclosure, query), \textit{4. Retention}, which is a description of where the result is stored and for how long and \textit{5. Recipients} who are the entities that can access the result of the operation.

As this task resembles answering explanatory linguistic queries, we employ a decoder-based model, which is better suited for natural language understanding and structured generation. Empirically, we found encoder-only architectures to be less effective for fine-grained extraction, as embedding-level noise degrades the reliability of centroid-based clustering across the 23 classes. This observation is consistent with prior findings \cite{gao2023retrieval}, which highlight limitations of embedding representations for fine-grained semantic discrimination.
We use a Llama3.1-8B-instruct to word-by-word extract these elements from privacy paragraphs and allow empty elements, if necessary, with the exception of \textit{data} type.

\subsection{Data Item/Purpose Keyword Mapping }

This stage identifies the most appropriate category for a given \{data, purpose\} pair decoded previously. We use a Llama3.1-8B-Instruct to map a batch of data items to a pre-assigned set of 23 keywords (22 keywords depicted in Fig.~\ref{fig:terminology}, with a detailed keyword list also in Appendix~\ref{subsec:appendix_DS}) and to map a batch of data purposes to a pre-assigned set of eight keywords. All the keywords were selected based the terms defined in DS declarations. 
We allow three generalised keywords for data-items: `other personal information', `generic information' and `negatives' (i.e., 23rd keyword - the text segment is not suitable as a data item - to filter noisy inputs decoded previously), and `other' keyword for data-purpose. 

The batched keyword mapping expects $N$ inputs to be orderly mapped to $N$ output keywords. 
$N=20$ is selected empirically to speed up keyword mapping (i.e., due to prompt overhead $~\sim100+$ words in decoding; {\bf cf.} Fig.~\ref{fig_appendix:data_item_mapping} and  \ref{fig_appendix:data_purpose_mapping} in the Appendix) while minimising errors. The errors are twofold. First, any output $\neq N$ produces an error as the model has not produced a one-to-one mapping. Second, hallucinations can occur when the decoder makes up keywords instead of selecting from the pre-assigned list. To avoid such errors, we train an encoder-based self-supervised verifier to be trained on successfully keyword-mapped outputs. The intuition of this is to have a lightweight encoder classifier trained on the decoded outputs of a larger decoder model. An encoder guarantees that each input can be one-to-one mapped to an appropriate class label without hallucinations. We train two such models for the two keyword-mapping tasks \textcolor{black}{and the training data from decoder models are synthetic in nature and are treated as pseudo-labels in self-supervised paradigm.} 

Based on these outputs, we create two matrices for each privacy policy, one for data collection and one for data sharing ({\bf cf.} Fig.~\ref{fig:first_example}). Each row of the matrix represents a data item keyword and each column represents a purpose. Multiple mentions of a single data-purpose pair increases the frequency. 
\textcolor{black}{Comparable data matrices are generated for the DS and the sanitisation steps we perform for this are explained in Appendix~\ref{subsec:appendix_DS_sanitisation}}.  

\subsection{Dataset}
We analyse 3,400 unique privacy policies linked to 7,770 top-ranked (based on download count) mobile games on Google Play, each with over 1M installs, including 174 titles exceeding 100M downloads. The dataset was constructed via a series of preprocessing steps, including language filtering, format validation, and file size constraints. We observed a policy reuse rate of 64.9\% with 1,077 policies covering two to ten games each. Within the tail we observed 21 policies, each representing more than 20 games. A detailed description is available in the Appendix~\ref{subsec:appendix_dataset}. For the generalisation of our results to non-game apps, we analyse 1,254 unique privacy policies representing 1,711 apps.

\section{Metrics for Compliance Quantification}
\label{sec:metrics}

\subsection{Data Practice Compliance}

When developers self-declare their \emph{data practices}—i.e., any activity involving first-party data collection or third-party data sharing—in the privacy policy (PP) and in the Data Safety (DS) section, it becomes essential to assess the consistency between the two sources. To this end, we define two metrics: PP compliance and DS compliance.

\textbf{PP compliance} quantifies the extent to which data items listed in the DS declaration are also explicitly mentioned in the relavent PP. Formally, we define it as the proportion of declared DS data items that can be verifiably found in the PP.

$$PP~compliance = n_{(PP~\cap~DS)}/n_{DS}$$

\begin{footnotesize}
    $$n_{\alpha}=\sum_{j\in J}f(d[j])~where~f(d[j])=1~if~d[j]\in{\alpha}~else~0$$
\end{footnotesize}

Here, $j\in J$ indicates the data item $d[j]$ belonging to the list of data items $J$ and $len(J)=23$ in this experimental setup. As a given data item $i$ could follow the high-level classification of $dp_i=c_0$ (i.e., first party data collection) or $dp_i=c_1$ (i.e., third party sharing), we calculate $PP~compliance$ separately for each data practice classification type. 

\textbf{DS compliance} measures the extent to which data items mentioned in the PP are also reflected in the DS declaration. Formally, it is defined as the proportion of data items identified in the privacy policy that are declared in the DS section.

$$DS~compliance = n_{(PP~\cap~DS)}/n_{PP}$$ 

Results of the PP and DS compliance are discussed in Sub.Sec.~\ref{subsec:data_practice_compliance}.

\subsection{Data Purpose Compliance}
\label{subsec:data_purpose_compliance}

For a given $\{PP, DS\}$ pair, we could observe the purpose compliance by comparing the presence of each individual purpose $k$ when a given data item $j$ is agreed as collected or shared among the pair. Therefore, we present our results per data item category and when the PP and DS do not agree on a data item, we specifically do not elaborate such results as PP compliance and DS compliance metrics already characterise them. We measure data purpose compliance as the intersection over union (IoU) of occurrences.

$$Purpose~compliance(k,j) = N_{(PP~\cap~DS)}/N_{(PP~\cup~DS)}$$

$$N_\beta = \#~p[k] \in \beta ~when~d[j]\in(PP~\cap~DS)$$

Results of the data purpose compliance are discussed in Sub.Sec. \ref{subsec:data_purpose_compliance}. The next section benchmarks PrivPRISM framework and highlights robust performance. 

\section{Benchmarking PrivPRISM }
\label{Sec:benchmarking}

As PrivPRISM is sequential; full ablation by module removal is not meaningful because later stages depend on structured outputs from earlier ones. We therefore report module-level ablations where evaluation is well-defined.

\begin{table}[t]
    \centering
    \renewcommand{\arraystretch}{1.1}
    \begin{tabular}{l |l l  | l l | l l}
        \hline
         & \multicolumn{2}{c|}{Ours (EDPC)} & \multicolumn{2}{c|}{F.T. Llama3.1} & \multicolumn{2}{c}{Z.S.GPT4o} \\
        Class & Pr & Re & Pr & Re & Pr & Re \\
        \hline
        $c_0$  & 0.94 & 0.74    & 0.90 & 0.21   & 0.88 & 0.69\\
        $c_1$  & 0.96 & 0.66    & 0.88 & 0.25   & 0.85 & 0.68\\
        $c_4$  & 0.73 & 0.41    & 0.51 & 0.19   & 0.87 & 0.54 \\
        \hline
        $\mu\bar{c}$ & 0.91 & 0.56 & 0.41 & 0.24 & 0.85 & 0.54\\
        \hline
        \multicolumn{7}{c}{$c_0$-first party collection, $c_1$-third party sharing}\\
        \multicolumn{7}{c}{$c_4$-introductory/generic, $\mu\bar{c}$-micro average for all}\\       
        \hline
    \end{tabular}
    \caption{Benchmarked test-set results for OPP-115 dataset conducted for our Explained Data Practice Classification (EDPC) module compared against fine-tuned Llama3.1 and zero-shot prompted GPT4o; both performing as explainable classifiers.  }
    \label{tab:explained_classifier}
\end{table}

\begin{figure}[t]
    \centering
    \includegraphics[width=0.45\textwidth]{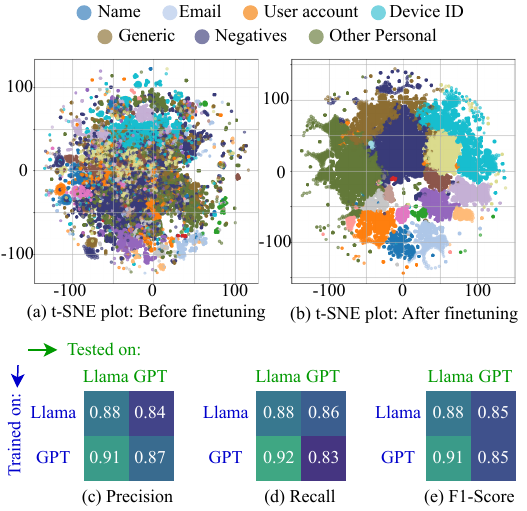}
    \caption{Data item mapping verifier embedding alignment and transferability}
    \label{fig:t-sne_prf1}
\end{figure}

\subsection{Explainable High-level Paragraph Classification}
\label{subsec:benchmark_exp_cls}

We benchmark the test set results of the OPP-115 dataset for the explained data practice classifier of our framework against fine-tuned Llama3.1 and zero-shot prompted GPT4o (via API access); in Tab.~\ref{tab:explained_classifier}. In this setup, both fine-tuned Llama3.1 and GPT4o operate strictly on explained classification setting which is comparable to our method's outputs.
We forward readers to Appendix Sub.Sec.~\ref{subsec:appendix_exp_cls} for full results for all classification classes and detailed explanations about the benchmarking setup. When classifying data collection ($c_o$), our method has a 4 and 6 percentage points better precision than Llama3.1 and GPT4o, respectively and 8 and 11 \textcolor{black}{percentage points} better in third-party sharing ($c_1$) classification. 
Micro average precision for all classes showcases 6 \textcolor{black}{percentage points}  better accuracy against GPT4o and 50 \textcolor{black}{percentage points} 
than Llama3.1 (due to poor performance of Llama in minority classes).
The OPP-115 dataset allows paragraphs to have multiple labels, however our method (and baselines discussed here) selects the most appropriate label, which reduces recall in benchmark results but prioritises precision for compliance tasks. To further validate the results are consistent with real world policies, we randomly (10 policies per 10KB increments of the file sizes) selected 50 policies with our framework’s classification results and manually verified the outputs. We observed first party collection labels are 95.3\% accurate while recalling 90.9\% instances and third-party sharing labels are 88.6\% accurate while recalling 95.4\% instances.

With respect to the explainability, we follow the same definition as in \cite{silva2024entailment}, where we measure the overlap percentage of a generated reasoning text against legal expert annotations for a given paragraph. Our method recorded an average 62.85\% overlap compared to 52.35\% of GPT4o and 57.77\% of Llama3.1. 

\subsection{Keyword Mapping and Self-supervised Mapping Verifier}
We select 106,971 data items and 61,877 purposes decoded for 1,000 privacy policies to then benchmark keyword mapping and verifier training. Llama3.1 resulted in 43.7\% error rate in data item one-to-one matching and 1.53\% hallucinations (e.g., made up class labels such as `cookies', `virtual items'). We ran the same mapping experiment using GPT4o, and the error rate was 22.3\% with 0.53\% hallucinations. Despite GPT4o being better, both of these models struggled with this seemingly straightforward task as an element-wise classification problem. Data item self-supervised verifier is trained to produce a 23-dimensional one-hot encoded output and is trained on the training set of either Llama3.1 (53,021) or GPT4o (74,394) data belonging to 860 policies. Fig.~\ref{fig:t-sne_prf1}(a) and (b) are illustrative examples of how encoder embeddings belonging to similar data items are aligned with Llama3.1-based training. More importantly, by conducting a transferability evaluation, where micro average precision, recall and F1 results shown in Fig.~\ref{fig:t-sne_prf1}(c-e), we observed that, irrespective of the method we used to create the training set (i.e., mappings generated by GPT4o or Llama3.1), we get similar F1 scores meaning the smaller encoder has captured the keyword mapping task well. The best advantage of the verifier is the ability to deploy in an inference setting and provide accurate mappings to decoder errors, even among the training dataset (as errors were not used for training). For full transferability results, refer to Appendix Tab.~\ref{tab:appendix_overall_results}.

Equipped with this knowledge, we performed the data purpose mapping and the respective self-supervised verifier training, with the exception of training only with Llama3.1-based mapping outputs. Initial error rate for decoder keyword mapping was 6.2\% and the hallucination rate was 0.6\%. We selected the training and testing datasets similar to the previous setup and
observed micro averaged 0.93 precision, 0.92 recall and 0.93 F1 score and concluded that the finetuned verifier module's outputs are in alignment with the larger 8B parameter decoder model. Next, we used this verifier module to correct the training data errors, similar to before and deployed it in PrivPRISM. Detailed alignment results are provided in Appendix Tab.~\ref{tab:purpose_map}.

\section{Results}
\label{sec:results}

In this section, we first conduct a privacy policy completeness analysis. We then evaluate data practice compliance for mobile games, as they are less likely to share privacy policies with web-based counterparts (e.g., a single policy covering both an app and its website) and are more frequently downloaded and used by minors. Next, we apply the PrivPRISM pipeline to non-game applications to assess category-wise compliance and compare these findings with those observed in games. Finally, we present an in-depth analysis of a selected subset of apps, highlighting privacy concerns associated with non-compliance.

\subsection{Policy Completeness}

\begin{figure}[t]
    \centering
    \includegraphics[width=0.48\textwidth]{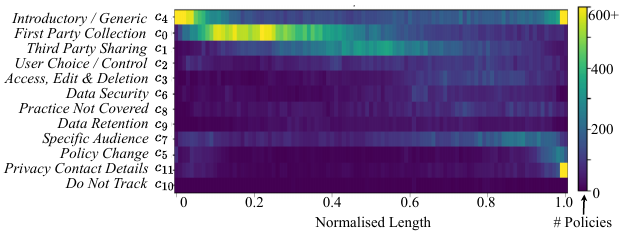}
    \caption{\textcolor{black}{Structural composition of a privacy policy - A column represents the data practice category distribution of the policy dataset at a given normalised length.} }
    \label{fig:completeness}
\end{figure}

Analysing paragraph-level classifications across 3,400 policies using the \textit{explained classifier} in the PrivPRISM framework reveals the general structural composition of privacy policies. As shown in Fig.~\ref{fig:completeness}, policies typically begin with introductory sections and end with contact information ($\sim$15\%). First-party collection ($\sim$40\%) and third-party sharing ($\sim$20\%) dominate the core content. Completeness for these two categories remains high (91.6\% and 90.3\%), whereas $c_{8}$ (64.7\%), $c_3$ (55.8\%), $c_9$ (42.8\%), and $c_{11}$ (6.8\%) show limited coverage, highlighting sparse user-data-control disclosures. This limited coverage is significant because GDPR and similar regulatory frameworks mandate clear disclosures on user data rights and control mechanisms, meaning that omissions in these categories reflect substantive compliance risks.

\subsection{Data Practice Compliance}
\label{subsec:data_practice_compliance}

We evaluated 3,400 policies against each of their most downloaded game and we observed average compliance scores of 82.07\% PP data-collect, 23.75\% DS data-collect, 68.52\% PP data-share, and 20.90\% DS data-share. We can observe that developers tend to disclose collection practices better than sharing practices. 
Out of the 3,400 pairs, 46.7\% of them achieved  100\% PP and DS compliance scores, indicating the data items declared across them overlapped perfectly. However, note that this is irrespective of indicated purposes. 

\begin{table}[ht]
    \centering
    \renewcommand{\arraystretch}{1.2}
    \begin{tabular}{l l l l l}
        \hline
        Downloads: & \multicolumn{2}{c}{$\sim 5M+$} & \multicolumn{2}{c}{$\sim 1-5M$}\\
        & Collect & Share & Collect & Share\\
        \hline
        PP Compliance & 81.23 & 68.08 & 83.28 & 69.17 \\
        DS Compliance & 25.44 & 22.58 & 21.29 & 18.47\\
        \hline
    \end{tabular}
    \caption{Overall Data Practice Compliance (\%) Landscape}
    \label{tab:appendix_overall_results}
\end{table}

Tab.~\ref{tab:appendix_overall_results} showcases how the compliance scores differ according to their popularity. We could see that PP compliance is slightly better at less popular games than with the most popular ones. However, the DS compliance remained the other way around. We will further elaborate on this trend explaining Fig.~\ref{fig:compliance_trends}.

\textbf{Policy re-usage affecting compliance:} 
A significantly higher mean PP compliance than DS, confirms higher data disclosures in privacy policies than what is required / declared for each game app.
Having a unified privacy policy for all games by a developer justifies this score, at the expense of reducing comprehension and clarity for end-users. To understand its impact, we aggregated all individual data safety declarations into a single \emph{super-data-safety} entry per shared policy. This aggregation led to a modest DS compliance improvement of 9.26 percentage points for data collection and 6.63 for data sharing supporting the view that policy reuse inflates apparent discrepancies when assessed at the per-app level.
Policy reuse, when applied across multiple games with heterogeneous functionality, can also contribute to overly generic or `catch-all' disclosures (e.g., `we collect personal data to provide services'), leading to overstatement and reduced specificity. This, in turn, increases the likelihood of PP–DS inconsistencies and undermine transparency. 
It is also worth noting that even with super-data-safety declarations, many policies still disclosed more data types than actually declared in the Play Store that suggest continued misalignment. 

\subsection{Data Purpose Compliance}

\begin{table}[h]
\centering
\footnotesize
\begin{tabular}{rrrrrrrr}
\hline
 & \textbf{p[0]} & \textbf{p[1]} & \textbf{p[2]} & \textbf{p[3]} & \textbf{p[4]} & \textbf{p[5]} & \textbf{p[6]}  \\
\hline
d[0]  & 0.71 & 4.00 & 0.78 & 1.40 & 7.50 & 33.06 & 20.41 \\
d[1]  & 2.63 & 17.09 & 6.76 & 1.96 & 1.55 & 50.00 & 11.81 \\
d[2]  & 20.00 & 7.49 & 14.66 & 5.71 & 12.88 & 43.48 & 35.10  \\
d[4]  & 0.00 & 7.69 & 0.00 & 0.00 & 0.00 & 40.00 & 7.14  \\
d[8]  & 16.55 & 1.45 & 34.39 & 6.57 & 6.37 & 1.76 & 25.76 \\
d[9] & 26.46 & 2.05 & 7.11 & 7.83 & 9.33 & 1.95 & 17.19 \\
d[11] & 0.00 & 23.91 & 11.36 & 0.00 & 0.00 & 2.13 & 11.32 \\
d[12] & 0.00 & 2.17 & 2.13 & 0.00 & 7.50 & 10.81 & 35.29 \\
d[15] & 10.00 & 0.00 & 0.00 & 12.50 & 16.67 & 0.00 & 44.44 \\
d[17] & 47.52 & 1.65 & 7.95 & 8.23 & 7.84 & 0.86 & 35.87 \\
d[18] & 37.49 & 0.91 & 12.87 & 14.98 & 6.05 & 4.03 & 36.60 \\
d[21] & 50.98 & 4.76 & 3.85 & 3.23 & 18.60 & 11.11 & 34.48\\
\hline
\end{tabular}
\caption{Data purpose compliance (\%): Each row represents $k^{th}$ purpose compliance score for a given valid data item category $j$.}
\label{tab:data_purpose}
\end{table}

We present the data–purpose compliance scores in Tab.~\ref{tab:data_purpose} for twelve data types. As described in Sub.Sec.~\ref{subsec:data_practice_compliance}, a detected purpose contributes to the compliance score only when both the privacy policy (PP) and the data safety (DS) declaration agree for a given data item. Data items (i.e. rows) not shown in the table fall into categories where PP–DS agreement is limited or entirely absent. Later in the results section, we further examine the observed collection and sharing frequencies, highlighting that despite the wide range of data categories available in DS declarations, developers consistently over-declare in PPs and under-declare in DS forms, resulting in limited overlap between the two.

Across all categories, developers most consistently disclose the collection and sharing of data items for \emph{p[6] – App functionality}, indicating that functional-requirement-related purposes are the most reliably communicated. We also observe relatively clearer alignment for \emph{d[8] – financial data} when used for \emph{p[2] – security, compliance, and fraud prevention}, and for \emph{d[9] – location data} when used for \emph{p[0] – app analytics}. Additionally, \emph{d[0–2,4] – name, email, user account, and phone data} show moderate consistency when disclosed for \emph{p[5] – account management} across both PP and DS.

However, almost no cell exceeds a 50\% compliance score, underscoring a major gap in how developers articulate why specific data types are collected or shared. For example, we can only observe 7.83\% purpose compliance for \emph{location data collected or shared for advertising}, despite this data item category observed in top-5 frequently declared in the privacy policies (cf. Fig.~\ref{fig:apk_evidence}). It is important to note that DS declarations naturally impose a fine-grained taxonomy that mobile app developers are expected to consistently reflect in their privacy policies. When such alignment is missing, users are ultimately presented with vague or overly broad descriptions of data practices, leading to blanket consenting and undermining meaningful transparency.

\subsection{Code-level Evidence Based Compliance}
\label{subsec:APK_evidence}

\begin{figure}[t]
    \centering
    \includegraphics[width=0.47\textwidth]{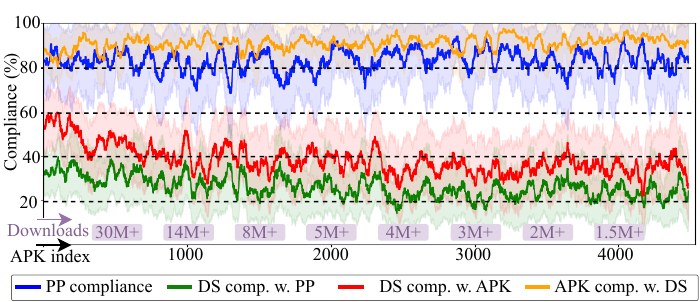}
    \caption{Four types of moving average filtered data item compliance scores based on the app popularity. Moving average window = 50, Shade = 0.5*STD} 
    \label{fig:compliance_trends}
\end{figure}

\begin{figure}[t]
    \centering
    \includegraphics[width=0.47\textwidth]{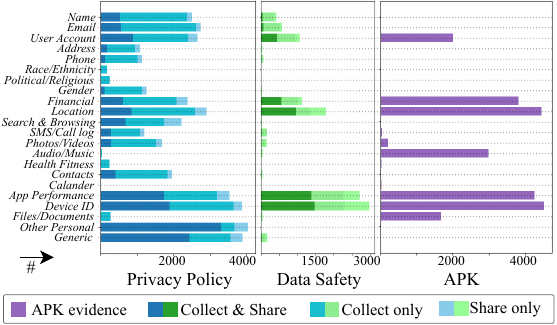}
    \caption{Number of data item declarations mentioned in PP, DS and within APK evidence for 4,538 game apps.} 
    \label{fig:apk_evidence}
\end{figure}

To examine how inconsistencies between PP and DS declarations manifest in source code based potential app behaviour, we analysed $\sim$5,000 game Android package files (APKs). Our analysis considered two complementary aspects: (1) the explicitly declared permissions requested in the manifest files, and (2) the sensitive dataflows inferred from method calls referenced in the compiled classes, both linked to specific data items accessed by the apps. Further details are included in Appendix~\ref{subsec:appendix_APK}.

Fig.~\ref{fig:compliance_trends} and Fig.~\ref{fig:apk_evidence} highlight three key findings. (1) APK-level evidence compliance w.r.t. DS declarations remains high and even surpasses PP compliance, suggesting that the data types disclosed in DS are generally reflected in both policy text and app behaviour. (2) However, the converse is not observed!, with many data items declared in PPs absent in DS and not evidenced in APKs, indicating that developers tend to include broad, precautionary disclosures to secure blanket user consent and reduce future liability rather than to mirror actual data practices. (3) Most game APKs access financial (84.3\%), location (98.3\%), and user account (58.7\%) information, yet these categories are only partially acknowledged in PPs (52.6\%, 64.1\%, 44.5\%) and are concerningly under-represented in DS declarations (<40\%). In contrast, device identifiers and analytics data show stronger alignment across PP, DS, and APKs—likely due to tighter regulatory scrutiny surrounding unique identifiers.

\subsection{Results for Non-Game Apps}
\label{subsec:appendix_generalisation_non_game}

We deployed our framework to analyse 1,254 unique privacy policies of non-game apps. Collectively, they represent 1,711 apps. Similar to the methodology explained in main text, we select the most popular (most downloaded) app for each unique privacy policy for compliance score calculation. The selection of unique privacy policies was based on the app category (e.g. Social, Communication, Tools, Medical, etc.) of their most popular non-game app and we selected 50 per each category when available.

\subsubsection{Game versus non-game data practices comparison}
\label{subsubsec:appendix_non_game_data_items}

\begin{figure}[t]
    \centering
    \includegraphics[width=0.47\textwidth]{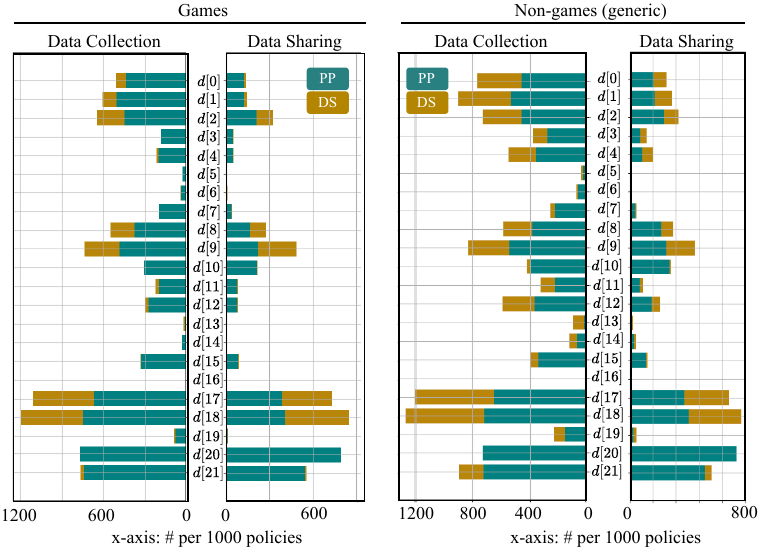}
    \caption{Data item collection or sharing frequency for game and non-game apps if mentioned in privacy policy (PP) or data safety (DS). X axis is normalised for easier comparison; i.e., the x-axis is normalised to show how many data items are observed per 1,000 privacy policies.}
    \label{fig:appendix_data_item_distro}
\end{figure}

For each PP and DS pair, our framework identifies which data items are collected and which data items are shared. While observing this for all 3,400 game related and 1,254 non-game related  unique privacy policies, we observed that some data items are more frequent and some rarely come up. To better understand results, we aggregated all the data item collection/sharing results and show the findings in Fig.~\ref{fig:appendix_data_item_distro}. Please note that, even if the privacy policy contains multiple instances about a single data item, we only consider one of those instances (low-level purpose classification omitted). Therefore, a data item can at most occur with the value 3,400 for games and 1,254 for non-games. For the easiness of comparisons, we normalise the results such that we show the collection or sharing numbers per 1,000 app-policy pairs. for a given data item, we stack the total number of declarations in DS and PP both to a single box plot and are emphasised using two colours; teal for PP and dark golden for DS.

\begin{figure}[t]
    \centering
    \includegraphics[width=0.47\textwidth]{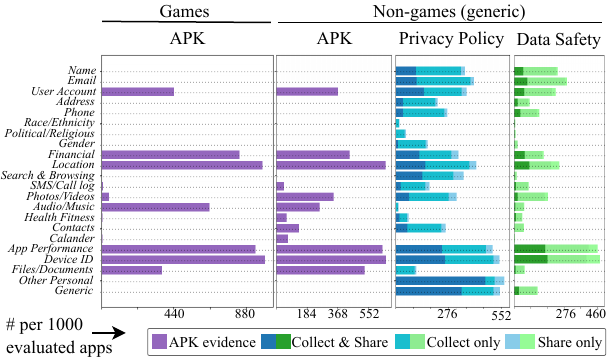}
    \caption{APK evidence analysis for non-game apps. Note: the x-axes are normalised for easier observations, therefore are in the same scale.}
    \label{fig:appendix_nongame_apk_evidence}
\end{figure}

\begin{figure*}[t]
    \centering
    \includegraphics[width=0.99\textwidth]{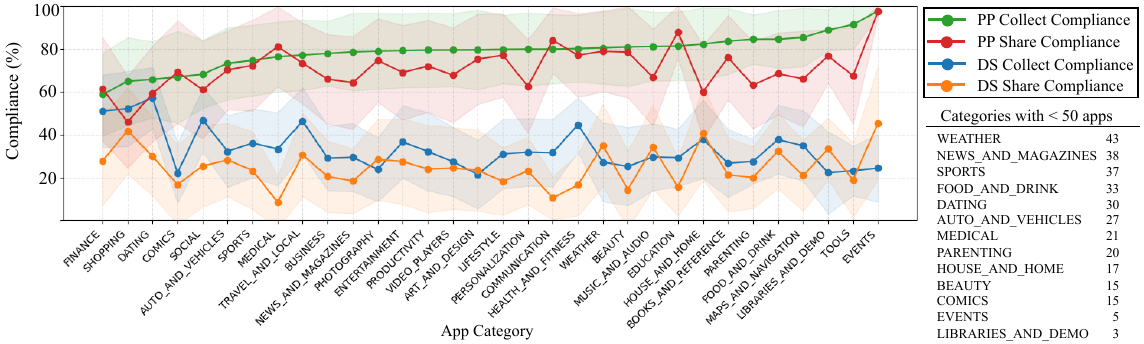}
    \caption{Privacy Policy and Data Safety collection and sharing compliance scores with respect to the corresponding non-game app's category. Note that several categories contained less than 50 apps when we selected the most popular non-game app corpus. Data points: mean compliance per category, Shade: 0.5*STD }
    \label{fig:appendix_app_category_distro}
\end{figure*}

We can deduce two main findings by observing the plots. First, the collection and sharing distributions for both game and non-game apps are similar with collection numbers generally higher than the sharing numbers. Second, for both games and non-games, the disclosures obtained via PPs are significantly higher than the DS, and is consistent with the main text's finding on higher PP compliance than the DS compliance. 

According to the results, DS declarations and PPs rarely declare about $d[5,6,13,14,16]$ categories \emph{(race / ethnicity / political / religious / photos / videos / audio / music / calendar)}. Classes $d[3,4,7,10,12,15]$ \emph{(address / phone / gender / search and browsing history / SMS / Messages / Call log / contacts)} are almost never declared in data safety declarations (dark golden colour) of games but more frequently found in non-game context. This highlights that game privacy policies are not really optimised for game specific data items, rather has a tendency to follow generic privacy policy formats, proving PPs are unlikely to align with end-user comprehension.

\subsubsection{Games versus non-games compliance score comparison}
\label{subsubsec:appendix_non_game_compliance}

\textcolor{black}{Among non-game app-policy pairs, 38.5\% demonstrated perfect alignment (compared to 46.7\% in games) and PP data-collect compliance score declined to 78.2\% (-3.9 percentage points), whereas the PP data-share score showed slight improvement (+1.5 percentage points) relative to games. App category wise compliance scores are depicted in the Fig.~\ref{fig:appendix_app_category_distro}. 
Medical and communication app categories demonstrated the lowest DS share compliance scores despite high PP collection and sharing scores, where it is likely that sensitive or high-risk data items are described broadly in privacy policies. This suggests that developers in these domains may prioritise broad legal coverage over accurate disclosure, leading to misalignment between stated and actual data practices and potentially confusing end-users.
Highest overall compliance scores were observed across the food and travel categories, representing apps that are frequently used by end-users (we omitted event category as the sample size was low). Lowest overall scores were observed by comics and news categories. Highest to lowest category scores observed a difference of nearly 23 percentage points.} 

Policy structure analysis shows that both games and non-games consistently disclose first-party collection (~91\%) and third-party sharing (~87–90\%). However, coverage of data security and retention is lower, with games lagging behind non-games (73.5\% vs. 78.2\% for security; 42.8\% vs. 55.1\% for retention.

\subsubsection{APK evidence analysis for non-game apps}
\label{subsubsec:appendix_apk_evidence_nongame}

We conducted APK evidence analysis for non-game apps, following the same procedure described in Sub.Sec.~\ref{subsec:APK_evidence}, with results shown in Fig.~\ref{fig:appendix_nongame_apk_evidence}. Financial and location data collection and sharing continue to show notable under-declaration patterns similar to game apps, whereas user account–related disclosures are comparatively more consistent.

In contrast, non-game apps exhibit a broader range of evidence for categories such as \emph{photos/videos, health/fitness, contacts, calendar, and files/documents}—an expected trend given their functionality. However, these categories often lack explicit justification, especially in DS disclosures, , suggesting that developers rely on generic statements rather than app-specific purposes. This highlights that while functional diversity in non-games leads to wider data access, transparency around the rationale for such access remains limited, posing user-awareness challenges.

\begin{figure*}[t]
    \centering
    \includegraphics[width=0.99\textwidth]{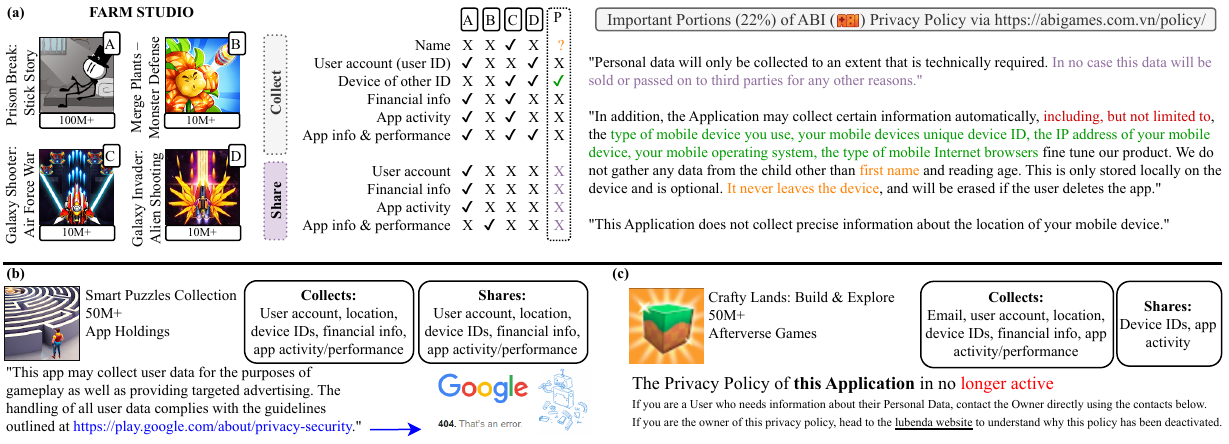}
    \caption{(a): A case study of `FARM STUDIO', (b,c): Existing policy links but invalid contents. Validated 2025.12 }
    \label{fig:pred_in_wild}
\end{figure*}

\subsection{When Games Break Rules!}

From a manual audit of the 50 lowest-compliance games identifed by PrivPRISM we uncover six recurring issues. (1) Several developers provided mismatched (e.g. a distant game developer policy provided instead of actual game policy) or placeholder privacy policy links, violating Google Play’s guidelines. (2) Popular games shared data with third parties without disclosure, while some omitted financial and performance data in their policies (e.g. in Fig.~\ref{fig:pred_in_wild}(a)) (3) Policies were often ambiguous, e.g., claiming names and ages were ``collected'' but never left the device, and (4) obscuring the implications of collecting IP addresses for approximate location. (5) In 38\% of cases, policy URLs redirected to layered sites, or inactive pages, frustrating access to the true policy. (6) Some developers reused identical policies across dozens of games under different URLs, complicating attribution and large-scale audits. Overall, every audited game showed discrepancies, with policies disclosing on average three more data items than their DS declarations, and ~33\% showing mismatches between developer and policy identities. We will elaborate more on specific findings in the next subsection.

\subsection{In-Depth Evaluation of the Manual Audit}
\label{subsec:appendix_manual_audit}

Among the 3,400 game policies we evaluated, we selected the 50 lowest privacy-compliant games (based on mean PP and DS compliant scores) and conducted a targeted manual evaluation to understand the implications. First, we have selected a developer as a case study, and we only discuss publicly available information as of Dec. 2025. 

\subsubsection{Case study: FARM STUDIO}
\label{subsubsec:appendix_farm_studio}

Fig.~\ref{fig:pred_in_wild}(a) presents four games by this developer, accompanied by respective DS sections and the important portions of the PP. Game A with the highest number of downloads (100M+) resulted in 0\% PP and DS compliance scores in PrivPRISM framework, and we also observed that the developer name and privacy policy names do not match; possibly due to citing game developer policy as their own.
The developer's website was simply a placeholder URL showing only `Hello World!', and their GMAIL address offered no further insight. \emph{01. It is against Google Play guidelines to provide a mismatched policy}.

Two of the games (A,B) share data with third parties without disclosure in PP, and the remaining two games had mismatched data collection practices, with some important details missing in the policy, like financial and app performance information collection. \emph{02. This is a data collection and sharing disclosure contradiction.} 

Additionally, parts of the PP were highly ambiguous. It indicates the developer gathering first name and age, only to later claim such information never leaves the device. Aiding the contradiction, Game C indicated collecting names in the DS, raising questions about \emph{03. highly ambiguous and contradicting policy text}. Furthermore, the PP mentions not collecting precise location but was unclear about the purpose of collecting IP addresses. \emph{04. IP address collection via internet connections may be used to detect end-user approximate location, where many privacy policies lack clarity}.

\subsubsection{Case study: BERNI MOBILE}

\textcolor{black}{With their most popular games downloaded more than 10M times, Berni Mobile’s privacy policy simply states that \emph{“Berni Mobile do not collect any personal user data”} and that \emph{“ad networks may access your unique device identifier through their own technologies and use it to target advertising to you.”} However, upon evaluating their popular title \emph{Cruciverba Italiano} (1M+ downloads), we found that the app collects and shares device or other IDs, approximate location, app info and performance data, and app activity for various purposes. Since approximate location can constitute personal data when linked to a unique device identifier, this represents a direct policy contradiction. Nevertheless, the fact that such a popular developer maintains an overly vague privacy policy of only about 100 words highlights a concerning lack of transparency and accountability in user data handling practices.}

\subsubsection{More ways to non-comply?}
\label{subsubsec:appendix_more_non_comply}

Fig.~\ref{fig:pred_in_wild}(b,c) portrays when policy URLs redirected users to active websites but with invalid privacy policies. The developer in (b) used a Google guideline (which is now invalid) as the privacy policy despite 50M+ downloads, and much sensitive information was collected and shared. This is a clear evasion of compliance; \emph{05. Google Play requires developers to submit an active privacy policy link. However, this requirement can be exploited with links containing privacy policy look-alike text.} Furthermore, 38\% of privacy policy links in the manual evaluation required us to read and click several other links to get to the actual privacy policy (hence the original URL contents were flagged as non-compliant with no data). Some of these contained language selections, region selections or were generic websites where users need to find true policy. 

Additionally, we found a developer named `Play Cool Zombie Sport Games' with eight games in this non-compliant list (41 active games in total) with customised and unique policy links for each of their games, only to discover after one level of redirection, lands on the same privacy policy belonging to `TEN SQUARE GAMES'. We were unable to verify any affiliation between the developer and the policy owner. \emph{06. Reusing identical policy content under different URLs may be a deliberate tactic to evade automated detection and complicate compliance audits at scale.} 

Manual verification of the remaining 62\% revealed discrepancies in every case, with privacy policies disclosing, on average, three more data items including location and financial data than their DS declarations. $\sim33\%$ showed mismatched developer and policy names, and some policies denied data collection while DS declared otherwise.
\section{Concluding Remarks}
\label{sec:conclusion}

We introduce PrivPRISM, a novel framework for fine-granular extraction and analysis of data practices in mobile app privacy policies. Unlike off-the-shelf LLM approaches, PrivPRISM employs a structured pipeline, achieving 6\% higher precision in explainable data practice classification compared to state-of-the-art GPT baselines. To support reliable real-world deployment, it integrates self-supervised verification modules to reduce generative errors and we define novel metrics for observing compliance. Applying PrivPRISM to PPs from $\sim 10,000$ of the most-downloaded Google Play games and generic apps ($\sim 5,000$ unique PPs), we identify potential discrepancies in 53\% (games) to 61\% (non-games) of cases. Code-level evidence analysis shows widespread PP and DS under-declarations among potential sensitive data-item categories alongside numerous other categories lacking explicit justification for collection or sharing. A manual audit of the 50 least compliant game-apps qualitatively reveals PP-DS contradictions, placeholder URLs, vague or conflicting statements, and tactics that may obscure auditability—all posing clear threats to end-user privacy. 

Future work can extend our framework in several directions. First, while our analysis highlights substantial discrepancies between privacy policies, data safety declarations, and static APK evidence, a fully comprehensive assessment would require dynamic analysis to observe real user interactions, runtime data flows, and network behaviours—elements that static techniques and fuzzing alone cannot reliably capture \cite{ahmad2020stadart, qu2017dydroid, zimmeck2017automated}. Second, our study assumes direct accessibility of privacy policies via developer-provided URLs; however, some policies require navigating complex website structures or hidden links, suggesting the need for more intelligent crawling methods capable of handling dynamic and nested content. However, it is noteworthy to mention that, it is mandatory for developers to provide a direct, non-geo-fenced and public facing URL during the app submission process. Finally, discrepancies may also arise when the same app appears across multiple platforms or markets, where metadata varies and disclosure taxonomies differ (e.g., Android vs. iOS). Extending PrivPRISM to reconcile cross-market versions and platform-specific disclosures would enable a more holistic compliance assessment.\\

\emph{\textbf{Ethics Statement:} All data analysed in this study were publicly available artefacts, including privacy policies, Google Play Data Safety disclosures, and application installation files. As such, this research did not require institutional ethics approval.}

\bibliographystyle{plain}
\bibliography{PrivPRISM/references}

\appendix
\newpage
\section{Appendix}
\label{sec:appendix}

\begin{figure*}[ht]
    \centering
    \includegraphics[width=0.99\textwidth]{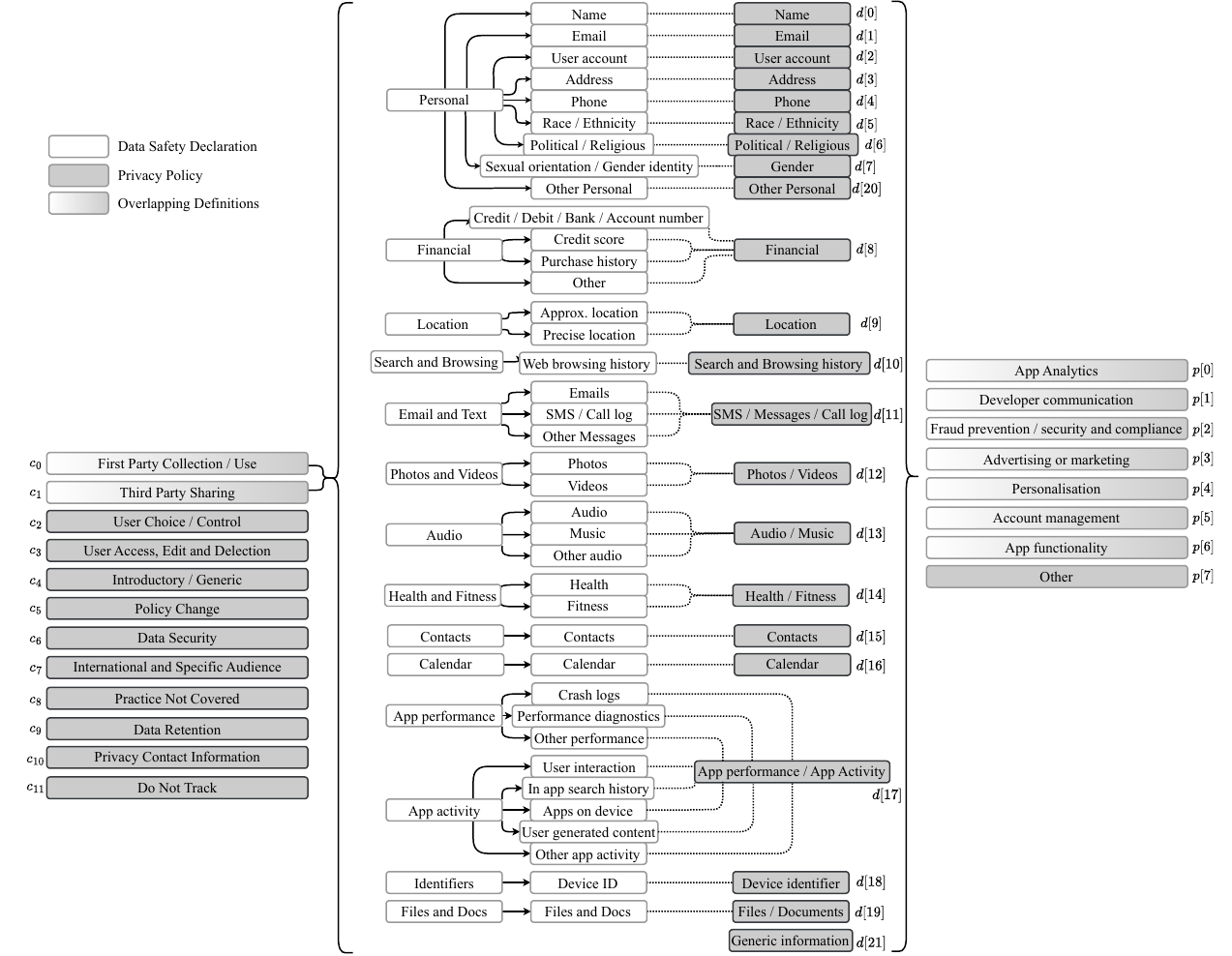}
    \caption{Combinations of Google Data Safety declaration labels, compared with  data practice, data item and data purpose categories we utilise in PrivPRISM framework.}
    \label{fig:all_combinations}
\end{figure*}

\subsection{Dataset}
\label{subsec:appendix_dataset}
Developers of top apps tend to place more emphasis on complete data safety declarations due to being more resourceful and concerns over public perception \cite{khandelwal2024unpacking}. Therefore, we narrow down our scope to consider top apps and, more specifically, games, as their privacy policies are less likely to be shared between corresponding web-based services. Based on a Google Play Store metadata crawl (approx 1.3M apps) we conducted during Q1 and Q2 of 2024, we select the top two percentile and filter out the games based on the number of download counts. 
We observed that among top apps, games and non-game apps are almost equally distributed, with non-game apps starting to dominate after the fifth percentile. During the crawl, we also downloaded the privacy policies via developer-declared URLs using the \texttt{PyWebCopy} Python library that saves the HTML file with linked resources such as images and JavaScripts. Based on the downloaded policies, we observed a 29.42\% failure rate due to errors in request refusals and incomplete downloads. Despite the mandatory requirement, 1.52\% did not contain a privacy policy link, 1.74\% contained PDF or TXT files as the privacy policy and 3.77\% of policies were hosted on the docs.google.com domain. We selected the successful 7,770 policies and respective games as our selection for analysis. The lowest download count was 1.3M, with 174 games with 100M+ and 28.7\% with 10M+ downloads. The average file size of the text-converted policies was 18.3KB (median 11KB), and we did not consider policies greater than 50KB ($<3\%$ of total) in length. 

Furthermore, despite the attempt to crawl the privacy policies via links provided with app metadata based on an English-speaking geo-location, we still observed 506 instances with non-English policies (40 Japanese, 59 Korean, 52 Portuguese, etc.), indicating that end-users are bound to use translation services to read these privacy policies. We also observed that 4,526 of these games shared 1,147 policy links; i.e., multiple games from the same developer sharing the same privacy policy. As an example, there were 67 games governed by voodoo.io privacy policy. After these basic sanitation, we were left with 3,400 unique privacy policies to be used for the experimental setup. Policy HTML to text conversion was done using the \texttt{BeautifulSoup4} Python library.

\subsection{Labels of Google Data Safety Declarations}
\label{subsec:appendix_DS}

In this appendix section, we provide a comprehensive overview of the labels used in Google's Data Safety declaration and how our fine-granular interpretations correlate with that. First, we observed all possible combinations of data practice declarations developers can provide for an app during the submission process. As illustrated in Fig.~\ref{fig:all_combinations}, data safety declaration starts with the data practice category, which is either data collection for first-party usage or data sharing with third parties. Also, the developers can disclose data security (request to delete) and whether the data is encrypted or not in transit from the mobile app to the collection party. When it comes to privacy policy, it is more fine-grained in classification. We follow common data practice types introduced by OPP-115 legal expert annotations, which are still widely used. These categories are shown in $c_0$ to $c_{11}$.

\begin{table*}[t]
    \centering
    \begin{footnotesize}
    \begin{tabular}{l l l l | l l l | l l l | l}
        \hline
         & \multicolumn{3}{c|}{PrivPRISM's EDPC} & \multicolumn{3}{c|}{F.T. Llama3.1} & \multicolumn{3}{c|}{Z.S. GPT4o}\\
        Class & Pr & Re & F1 & Pr & Re & F1 &  Pr & Re & F1 & Support\\
        \hline
        $c_0$ First Party Collection / Use          & 0.94 & 0.74 & 0.83 &  0.90 & 0.21 & 0.34 &    0.88 & 0.69 & 0.78 &      289\\
        $c_1$ Third Party Sharing / Collection      & 0.96 & 0.66 & 0.78 &  0.88 & 0.25 & 0.38 &    0.85 & 0.68 & 0.76 &      204\\
        $c_2$ User Choice / Control                 & 0.95 & 0.46 & 0.62 &  0.65 & 0.19 & 0.30 &    0.87 & 0.54 & 0.67 &      115\\
        $c_3$ User Access, Edit and Deletion        & 0.94 & 0.37 & 0.53 &  0.60 & 0.15 & 0.24 &    0.72 & 0.63 & 0.68 &      41\\
        $c_4$ Introductory / Generic                & 0.73 & 0.41 & 0.52 &  0.51 & 0.25 & 0.33 &    0.94 & 0.25 & 0.39 &      118\\
        $c_5$ Policy Change                         & 1.00 & 0.45 & 0.62 &  1.00 & 0.21 & 0.34 &    0.71 & 0.69 & 0.70 &      29\\
        $c_6$ Data Security                         & 0.92 & 0.55 & 0.69 &  0.79 & 0.24 & 0.37 &    0.86 & 0.60 & 0.70 &      62\\
        $c_7$ International \& Specific Audiences   & 0.91 & 0.72 & 0.80 &  0.89 & 0.28 & 0.43 &    0.80 & 0.73 & 0.77 &      60\\
        $c_8$ Practice not covered                  & 0.64 & 0.19 & 0.29 &  0.14 & 0.46 & 0.21 &    0.71 & 0.09 & 0.16 &      110\\
        $c_9$ Data Retention                        & 1.00 & 0.23 & 0.38 &  1.00 & 0.12 & 0.21 &    0.90 & 0.35 & 0.50 &      26\\
        $c_{10}$ Privacy Contact Information        & 0.97 & 0.66 & 0.79 &  0.79 & 0.20 & 0.31 &    0.93 & 0.45 & 0.60 &      56\\
        $c_{11}$ Do Not Track                       & 1.00 & 0.60 & 0.75 &  1.00 & 0.60 & 0.75 &    0.83 & 1.00 & 0.91 &      5\\
        \hline
        $\mu\bar{c}$ Micro Average                  & 0.91 & 0.56 & 0.69 &  0.41 & 0.24 & 0.31 &    0.85 & 0.54 & 0.66 &      1115\\
        $m\bar{c}$ Macro Average                    & 0.91 & 0.50 & 0.63 &  0.76 & 0.26 & 0.35 &    0.83 & 0.56 & 0.63 &      1115\\
        \hline
    \end{tabular}
    \end{footnotesize}
    \caption{Data Practice Classification Results. EDPC: Explained Data Practice Classifier, F.T.:Fine-tuned, Z.S.:Zero Shot}
    \label{tab:explained_classifier_appendix}
\end{table*}

Next, Google's Data Safety declaration provides detailed labels about what data categories (e.g., location data) and attributes (e.g., approximate or precise location) are shared or collected. These finer levels of detail are not provided for data encryption and deletion request declarations. When deciding the data items suitable to be identified from a privacy policy, we closely followed this structure and we decided on 22 data items $d[0]$ to $d[21]$. The final data item of `Generic Information' is introduced as we empirically observed many privacy policies using terms such as `we collect your information',  which are highly ambiguous, nonetheless, still require some attention. Also, it classifies terms that are out of domain here, such as `cookies'.  During the keyword mapping stage of our framework, we assigned an additional data item class $d[22]=negative$ that provides a decoder model to observe a text segment and decide as not suitable. This helps in error correction, where a sentence with multiple data items is separated using traditional NLP techniques. E.g. Word `etc.' after a list of data items needs to be classified as a negative if it was captured.

Lastly, the Data Safety Declaration provides details for each data category (not attribute level) about why this data category is collected or shared and can be one of seven data purposes as depicted in $p[0]$ to $p[7]$. App analytics and app functionality purposes are heavily biased towards mobile applications but other categories are generalised, therefore we followed the same structure in privacy policy based purpose identification. Additionally, we introduce the `other' category to classify any purpose that does not belong to one of these. In summary, this hierarchical structure allows us to have a fine-grained analysis of the privacy policies and to detect potential compliance violations in their respective mobile applications.

\subsection{Sanitisation of the Data Safety Declarations}
\label{subsec:appendix_DS_sanitisation}

\textcolor{black}{HTML files of the Data Safety (DS) pages were sanitised using a Document Object Model (DOM)–based approach with python library \texttt{BeautifulSoup} by referencing necessary division classes, which represents the page as a hierarchical tree of elements. This enabled the identification of primary data practice sections, descriptive statements, and nested categories detailing the types of data collected, their purposes, and any sharing practices. The extracted components were annotated with lightweight structural tags to retain semantic relationships while ensuring machine-readable consistency across all apps. For example, a disclosure such as “Data Shared → Device or other IDs → Data shared and for what purpose → Analytics, Advertising or marketing” (this is the same example shown in Fig.~\ref{fig:first_example}) is encoded as:}

\begin{itemize}
    \item \texttt{<d\_prac> Data Shared} 
    \item \texttt{<d\_cata> Device or other IDs} 
    \item \texttt{<d\_detl> Data shared and for what purpose} 
    \item\texttt{<d\_valu> Analytics, Advertising or marketing}
\end{itemize}

\subsection{APK Evidence Extraction}
\label{subsec:appendix_APK}

\textcolor{black}{To complement the analysis of developer-declared data practices in Privacy Policies and Data Safety declarations, we conducted static code analysis on the corresponding Android application packages (APKs). Each APK file was parsed using Androguard \cite{androguard}, an open-source reverse-engineering toolkit for Android that supports programmatic inspection of manifests, compiled code, and control/data flow graphs. The goal of this step was to extract code-level evidence of data access that could be aligned against the declared information.}

The analysis proceeded in two stages. First, the \emph{AndroidManifest.xml} of each APK was parsed to obtain all explicit permission requests. These permissions describe what categories of user or device data the application claims to access (e.g., \emph{ACCESS\_FINE\_LOCATION}, 
\emph{READ\_CONTACTS}).
Second, the compiled Dalvik bytecode (contained in the \texttt{classes.dex} files) was examined through Androguard’s \texttt{Analysis()} module. 
The resulting method analysis graph was traversed to identify API method invocations that are known to require or imply certain permissions. These API–permission associations were derived from a precompiled mapping resource (\texttt{api\_permission\_map}), which aligns Android framework API calls to the corresponding sensitive permissions defined by the Android SDK documentation.

\textcolor{black}{Each permission (either declared in the manifest or inferred from API usage) was then mapped to high-level data items. For example;
}

\begin{itemize}
    \footnotesize
    \item \texttt{android.permission.ACCESS\_FINE\_LOCATION → Location }
    \item \texttt{android.permission.ACCESS\_COARSE\_LOCATION → Location } 
    \item \texttt{android.permission.READ\_MEDIA\_IMAGES → photos/videos } 
\end{itemize}

\begin{table*}[ht]
    \centering
    \begin{footnotesize}
    \begin{tabular}{l|ccc|ccc|ccc|ccc|r }        
        \hline
        \vspace{-2ex} 
        & \multicolumn{3}{c|}{} 
        & \multicolumn{3}{c|}{} 
        & \multicolumn{3}{c|}{} 
        & \multicolumn{3}{c|}{} &  \\ 
             & \multicolumn{3}{c|}{Trained on: GPT } 
             & \multicolumn{3}{c|}{Trained on: GPT } 
             & \multicolumn{3}{c|}{Trained on: Llama } 
             & \multicolumn{3}{c|}{Trained on: Llama } &  \\

             & \multicolumn{3}{c|}{Tested on: GPT} 
             & \multicolumn{3}{c|}{Tested on: Llama} 
             & \multicolumn{3}{c|}{Tested on: Llama} 
             & \multicolumn{3}{c|}{Tested on: GPT} &  \\
        Class                               & P&R&F1             & P&R&F1             & P&R&F1             & P&R&F1             & Support  \\        
        \hline
        \vspace{-1ex} 
        & \multicolumn{3}{c|}{} 
        & \multicolumn{3}{c|}{} 
        & \multicolumn{3}{c|}{} 
        & \multicolumn{3}{c|}{} &  \\ 
        Name          &0.97 & 0.92   & 0.94 & 0.98 & 0.95 & 0.96 & 0.97 & 0.93 & 0.95  &0.96 & 0.92 & 0.94  & 199~149*\\
        Email         &0.99 & 0.98   & 0.99 & 0.99 & 0.99 & 0.99 & 0.97 & 0.98 & 0.97  &0.98 & 0.99 & 0.98  & 347~295*\\
        User Acct  &0.95 & 0.80 & 0.87  & 0.96 & 0.67 & 0.79 & 0.84 & 0.69 & 0.75 & 0.93 & 0.83 & 0.88  & 475~436*\\
        Financial     &0.95 & 0.98 & 0.96  & 0.94 & 0.91 & 0.92 & 0.94 & 0.85 & 0.89 & 0.94 & 0.99 & 0.97  & 270~205*\\
        Location      &0.98 & 0.94 & 0.96  & 0.93 & 0.95 & 0.94 & 0.90 & 0.92 & 0.91 & 0.96 & 0.95 & 0.95  & 220~364*\\
        App Perfo &0.90 & 0.93 & 0.91  & 0.85 & 0.95 & 0.90 & 0.91 & 0.78 & 0.84 & 0.81 & 0.97 & 0.88  & 529~402*\\
        Device ID     &0.87 & 0.96 & 0.91  & 0.93 & 0.95 & 0.94 & 0.88 & 0.92 & 0.90 & 0.88 & 0.95 & 0.91  & 744~487*\\
        Gen. Info&0.75 & 0.48 & 0.58  & 0.91 & 0.96 & 0.94 & 0.90 & 0.94 & 0.94 & 0.77 & 0.40 & 0.52  & 1005~2972*\\
        \hline
        \vspace{-1ex} 
        & \multicolumn{3}{c|}{} 
        & \multicolumn{3}{c|}{} 
        & \multicolumn{3}{c|}{} 
        & \multicolumn{3}{c|}{} &  \\ 
        Micro Avg &0.87 & 0.83 & 0.85  &0.91 & 0.92 & 0.91 &0.88 & 0.88 & 0.88  &0.84 & 0.86 & 0.85  &8271~6210*\\
        \hline
    \end{tabular}  
    \end{footnotesize}
    \caption{Transferability of self-supervised mapping verifier training} \vspace{-2mm}
    \label{tab:appendix_transfer_baselines}
\end{table*}

\subsection{Explained Paragraph Classification}
\label{subsec:appendix_exp_cls}

OPP-115 dataset introduced by \cite{opp115} contains 115 legal expert- annotated web-privacy policies, and we use the first 90 privacy policies for necessary training and 25 policies for testing and presenting the results. No hyperparameter tuning is done, and for GPT4o-related testing, the training dataset is not used. The original dataset consisted a high-level category called `other' and total of 10 high-level class labels, however, due to the ambiguity of `other' class \cite{privbert, tang2023policygpt, silva2024entailment}, we aggregate some lower-level class labels to create 12 total classes. They are represented in Fig.~\ref{fig:all_combinations} from $c_0$ to $c_{11}$. 54.5\% of the training paragraphs contain a single label per paragraph and with the expectation of identifying the most confident class label as a representation for the entire paragraph, we employ a multi-class-single-label classification strategy. We do this by observing the classification confidence distribution over all classes of our finetuned PrivBERT encoder. However, for observing the results, we still use the original labels for the readers to directly compare the results against literature. 
\textcolor{black}{This adaptation improves the precision by 6 and 2 percentage points for first-party collection and third party sharing compared to traditional PrivBERT multi-label setup. The recall drops by 11 and 17 percentage points and this behaviour is expected moving from multi-label to single-label configuration. In a manual review of 50 mobile game privacy policies, our method retained 90.9\% recall for first-party collection, and 95.4\% for third-party sharing, hence, confirming no-drop in recall performance in a real world deployment}. 
In contrast, \cite{tang2023policygpt} modified the precision, recall and F1 score calculations in favour of a single-label classification; however, we do not follow that strategy to boost results, and certain recall numbers in our results in Tab.~\ref{tab:explained_classifier_appendix} would be lower for this reason. 

For the PrivPRISM framework, we are mostly interested in the accuracy of $c_o$ and $c_1$ class predictions. Despite PolicyGPT claiming impressive results with GPT platform, we were unable to reproduce the results and we believe our traditional way of defining the metrics and the train-test selection may have affected that. To get GPT4o results, we follow a similar prompt as in PolicyGPT and is shown in Fig.~\ref{fig_appendix:gpt4-prompt}. We also compare our results against a fine-tuned Llama3.1 version using the training prompt strategy as shown in Fig.~\ref{fig:llama3-prompt}. We use parameter efficient fine tuning (PEFT) with low-rank adaptation (LoRA) for the training set over five epochs. Llama3.1 finetuning results show that performance for $c_4$ - Introductory/Generic and $c_8$ - Practice Not Covered is poor, reducing the average precision values. It also struggles with Recall values.

\subsection{Fine-granular Data Practice Decoding and Keyword Mapping}
\label{subsec:appendix_decoding}

Fig.~\ref{fig_appendix:fine_granular_prompt} shows the prompt we used for extracting the five elements (data, purpose, processing, storage, recipients) for a given policy text paragraph. The separate elements (e.g. comma separated) of the `data' field are then processed in a batched setting using the prompt we show in Fig.~\ref{fig_appendix:data_item_mapping} for data item keyword mapping. Outputs collected via this stage for 1,000 privacy policies are used as a training dataset (after filtering the decoding errors) for the self-supervised verifier module. 

Tab.~\ref{tab:appendix_transfer_baselines} shows the transferability results for the data item mapping verifier. The table shows results for eight data items that are frequently observed, however, we give the micro average precision, recall and F1 scores at the end for all 23 class labels. 

Similarly, Fig.~\ref{fig_appendix:data_purpose_mapping} shows the prompting we used for data purpose keyword mapping and Tab.~\ref{tab:purpose_map} show how the alignment output of the trained self-supervised data-purpose verifier module, compared against the original Llama3.1-8B-Instruct purpose mapping outputs.

\textcolor{black}{\textit{Note:} We have attempted to identify data items, purposes, recipients etc., by observing the contextualised embeddings generated by PrivBERT encoder model to an input token sequence. However, we did not observe a clear separation in high-dimensional vector space (e.g. distinct clusters) and the token window to consider was inconclusive. Instead, due to the llama3.1 instruct decoders empirical performance with explanatory linguistic questions such as “what data items present in this text?” provided straightforward answers.}

\begin{figure*}[t]
\centering
\begin{promptboxlisting}[GPT4o Zero Shot Prompt]
(*@\textcolor{brown}{I will give you the annotation scheme consisting of twelve data practice categories with explanations. The annotations are to the website's privacy policy.}@*)
(*@\textcolor{brown}{1. First Party Collection / Use: how and why a service provider collects user information.}@*)
(*@\textcolor{brown}{2. Third Party Sharing / Collection: how user information may be shared with or collected by third parties.}@*)
(*@\textcolor{brown}{3. User Choice / Control: choices and control options available to users.}@*)
(*@\textcolor{brown}{4. User Access, Edit and Deletion: if and how users may access, edit, or delete their information}@*)
(*@\textcolor{brown}{5. Introductory / Generic: if mentions about generic information or if its introductory}@*)
(*@\textcolor{brown}{6. Policy Change: if and how users will be informed about changes to the privacy policy.}@*)
(*@\textcolor{brown}{7. Data Security: how user information is protected.}@*)
(*@\textcolor{brown}{8. International and Specific Audiences: practices that pertain only to a specific group of users (e.g., children, Europeans, or California residents).}@*)
(*@\textcolor{brown}{9. Practice not covered: mentions that a practice is not covered by that privacy policy.}@*)
(*@\textcolor{brown}{10. Data Retention: how long user information is stored.}@*)
(*@\textcolor{brown}{11. Privacy contact information: how the users could contact for relevant information}@*)
(*@\textcolor{brown}{12. Do Not Track: if and how Do Not Track signals for online tracking and advertising are honored}@*)
(*@\textcolor{brown}{Based on the text provided, select the most matching category and provide a reason. The reason is a text excerpt (annotation) from the provided paragraph itself that explains best for the matching category provided.Please follow the output structure like this;
Matching category = 'category'
Reasoning = 'text annotation'}@*)
\end{promptboxlisting}
\caption{Prompt used in GPT4 Zero shot evaluation.}
\label{fig_appendix:gpt4-prompt}
\end{figure*}

\begin{table}[ht]
\centering
\begin{footnotesize}
\begin{tabular}{lcccc}
\hline
Class & Pr & Re & F1 & Support \\
\hline
0 - Analytics & 0.96 & 0.85 & 0.91 & 627 \\
1 - Dev. Commu. & 0.80 & 0.88 & 0.84 & 320 \\
2 - Fraud Prevention & 0.95 & 0.95 & 0.95 & 1530 \\
3 - Advertising & 0.97 & 0.95 & 0.96 & 590 \\
4 - Personalisation & 0.86 & 0.95 & 0.90 & 796 \\
5 - Account Manage & 0.96 & 0.88 & 0.92 & 847 \\
6 - App Functionality & 0.92 & 0.95 & 0.94 & 580 \\
7 - Other & 0.94 & 0.91 & 0.93 & 1030 \\
\hline
Micro Avg & 0.93 & 0.92 & 0.93 & 6320 \\
\hline
\\   
\end{tabular}
\end{footnotesize}
\caption{Alignment of 110M encoder model against the ground truth assumption from a 8B decoder model}
\label{tab:purpose_map}
\end{table}

\subsection{How Did We Manually Verify Non-compliant Games?}
\label{subsec:appendix_how_did_we_check}

\begin{figure}[ht]
    \centering
    \includegraphics[width=0.49\textwidth]{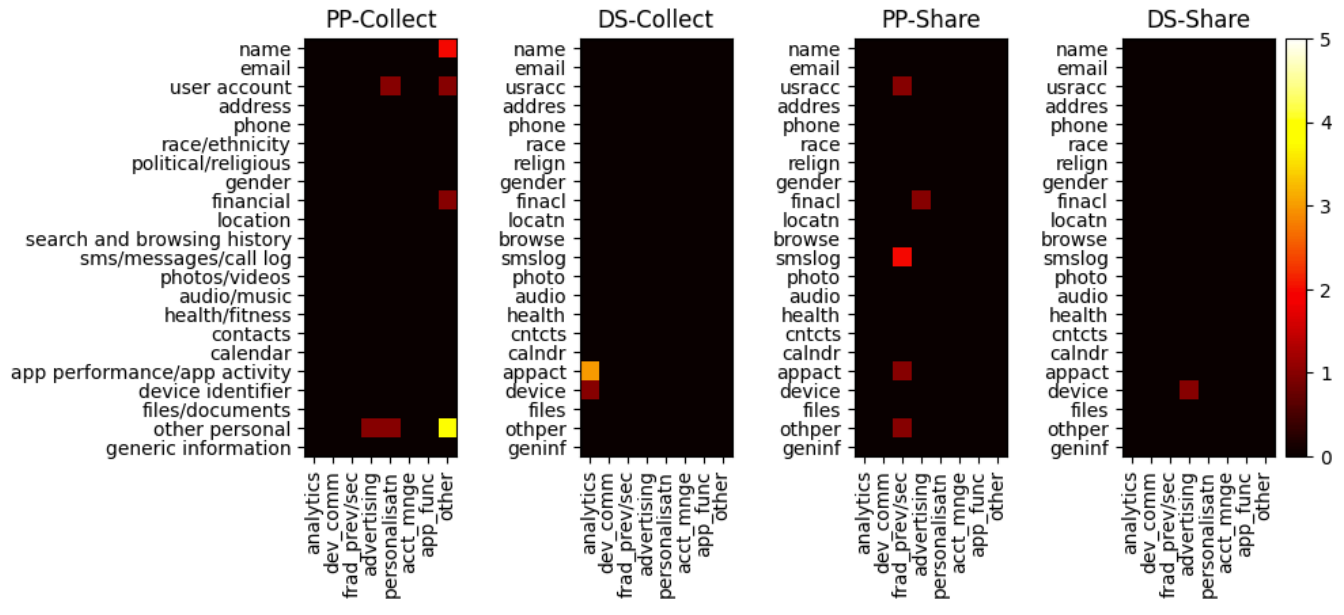}
    \caption{An example output from our framework for a given PP, DS pair with $0\%$ overall compliance score. Note: For output visualisation only - a more readable version is discussed in the main text, Fig.~\ref{fig:first_example}.}
    \label{fig:appendix_example}
\end{figure}

Fig.~\ref{fig:appendix_example} showcases an example output we receive from the framework when we query with a privacy policy and a data safety declaration. In this particular example, we did not see any alignment between the PP and DS (i.e. 0\% compliance). Next, we go through our decoded information; i.e., which heading, which paragraph mentions about which data types and purposes in the privacy policy. For this example, we observed the phrase \emph{You hereby further grant to [developer name] the unconditional, irrevocable right to use and exploit your name, username, display name, avatar and any other information} emphasising the privacy policy mentions the collection of such information, which was not observed in data safety. Also, we were unable to locate device identifiers and app performance activity information as depicted on the data safety declaration in the privacy policy. Having the automated framework assists in flagging such non-compliant pairs, showcasing where particular data item / purpose indications occur in the privacy policy,  and visually showcasing how the non-alignments of PP compare against DS and at which frequency.

\begin{figure}[ht]
\centering
\begin{promptboxlisting}[Llama3.1 8B Training Prompt]
DEFAULT_PROMPT = (*@\textcolor{brown}{"Below is a paragraph from a privacy policy, classify it according to the data practice selecting from the list below and provide a suitable reasoning extracted from the policy text."}@*)
DATA_PRACTICE_LIST = (*@\textcolor{brown}{"[\"First Party Collection / Use ... Do Not Track\"]"}@*)
training_prompt = f""" 
(*@\textcolor{brown}{\string#\string#\string# Instruction:}@*) {DEFAULT_PROMPT}
(*@\textcolor{brown}{\string#\string#\string# Data practice list:}@*) {DATA_PRACTICE_LIST}
(*@\textcolor{brown}{\string#\string#\string# Paragraph:}@*) {paragraph}
(*@\textcolor{brown}{\string#\string#\string# Class:}@*) {data_practice}
(*@\textcolor{brown}{\string#\string#\string# Reason: Because this paragraph mentions:}@*) {data_reason_annotation}"""
\end{promptboxlisting}
\caption{Prompt used in Llama3.1 training. Not indicated in brown are python variables which are self-explanatory.}
\label{fig:llama3-prompt}
\end{figure}

\begin{figure*}[ht]
\centering
\begin{promptboxlisting}[Llama3.1 Instruct-8B Inference Prompt 1]
messages = [
{"role": "user", "content": (*@\textcolor{brown}{"By going through the following privacy policy text, identify each suitable segment of text according to the following structured components:}@*)
(*@\textcolor{brown}{1.data: what type of data,}@*)
(*@\textcolor{brown}{2.purpose: why is this data required,}@*)
(*@\textcolor{brown}{3.processing: in which circumstances this data is utilized,}@*)
(*@\textcolor{brown}{4.storage: how long this data is stored or retained,}@*)
(*@\textcolor{brown}{5.recipients: who are the intended users for this data.}@*)
(*@\textcolor{brown}{Please note that sometimes it might not be possible to fill all five types above, if so, leave them empty. Provide the output as a JSON list."}@*)}
{"role": "user", "content": (*@\textcolor{brown}{"Here is an example:}@*)
(*@\textcolor{brown}{processed\_jsons = {[}\{'data': 'what type', 'purpose':'why', 'processing':'how utilized', 'storage':'how long stored', 'recipients':'who are recipients'\}, ...{]}"}@*)}
{"role": "user", "content": (*@\textcolor{brown}{"privacy policy text:"}@*) + pp_text} ]
\end{promptboxlisting}
\caption{Prompt used in Llama3.1 Instruct inference in the module for fine-granular detail extraction. Not indicated in brown are python variables which are self-explanatory.}
\label{fig_appendix:fine_granular_prompt}
\end{figure*}
\begin{figure*}[ht]
\centering
\begin{promptboxlisting}[Llama3.1 Instruct-8B Inference Prompt 2]
messages = [
{"role": "user", "content": (*@\textcolor{brown}{"The task is to map different data items in a given evaluation\_list with the most suitable keywords.}@*)
(*@\textcolor{brown}{Predefined keyword\_list: ['name', 'email', 'user account', 'address', 'phone', 'race/ethnicity', 'political/religious', 'gender', 'financial', 'location', 'search and browsing history', 'sms/messages/call log', 'photos/videos', 'audio/music', 'health/fitness', 'contacts', 'calendar', 'app performance/app activity', 'device identifier', 'files/documents', 'other personal'].}@*)
(*@\textcolor{brown}{Match each item in the following evaluation\_list to the most relevant keyword from the predefined keyword\_list above.}@*)
(*@\textcolor{brown}{Return the results in this format. output\_list=['data item1': 'keyword1', 'data item2': 'keyword2', ...]. Do not include explanations or extra text.}@*)
(*@\textcolor{brown}{if something in evaluation\_list is too generic, output it as 'generic information' but this output is discouraged}@*)
(*@\textcolor{brown}{Please use the 'N/A' to indicate if an item is not suitable (outlier) as a data practice item type."}@*)}
{"role": "user", "content": (*@\textcolor{brown}{"evaluation\_list = "}@*) + str(eval_set)} ]
\end{promptboxlisting}
\caption{Prompt used in Llama3.1 Instruct inference for data item keyword mapping. Not indicated in brown are python variables which are self-explanatory.}
\label{fig_appendix:data_item_mapping}
\end{figure*}
\begin{figure*}[ht]
\centering
\begin{promptboxlisting}[Llama3.1 Instruct-8B Inference Prompt 3]
messages = [
{"role": "user", "content": (*@\textcolor{brown}{"The task is to map the different purpose phrases in a given purpose\_list with the most suitable keywords.}@*)
(*@\textcolor{brown}{Predefined keyword\_list = ['analytics', 'developer communication', 'fraud prevention/security', 'advertising', 'personalization', 'account management','app functionality', 'other']}@*)
(*@\textcolor{brown}{Match each item in the following purpose\_list with the most relevant keyword from the predefined keyword\_list above}@*)
(*@\textcolor{brown}{Return the results in this format. output\_list = {'purpose item1':'keyword1', 'purpose item2':'keyword2',......}. Do not include explanations or extra text."}@*)}
{"role": "user", "content": (*@\textcolor{brown}{"purpose\_list ="}@*) + str(eval_set)} ]
\end{promptboxlisting}
\caption{Prompt used in Llama3.1 Instruct inference for data purpose keyword mapping. Not indicated in brown are python variables which are self-explanatory.}
\label{fig_appendix:data_purpose_mapping}
\end{figure*}

\emph{Discussion on usage:} The Main contribution of this work, by presenting the PrivPRISM framework, is that it can be utilised for automatic detection of discrepancies between Google Play data safety declarations and developer privacy policies. While some flagged discrepancies may stem from ambiguity or interpretation differences rather than intentional malpractice, especially due to highly ambiguous privacy policy text, we envision its adoption as a complementary aid to enhance transparency and accountability, ideally integrated into human-in-the-loop auditing workflows for robust, real-world impact.

\subsection{Actionable Insights} 
\label{subsec:appendix_actionable_insights}

\textcolor{black}{According to Google Play policy, it is mandatory for app developers to provide an active, public and non-geo fenced privacy policy URL in the app listings.  Non-compliance could lead to banned apps or banned developer profiles by Google. Such attempts will be identified during three stages;  (1) Google’s new app publishing phase, (2) For existing apps - during the regular app updating process and (3) For existing apps without valid privacy policy links – during Google's routine ecosystem checks.}

\textcolor{black}{Developers could potentially avoid all three of the stages above if there is an active, public and unrestricted page linked as a privacy policy, but the content could be vague, placeholder or boilerplate. PrivPRISM becomes a valuable tool in this context;}

\textcolor{black}{\textbf{01-	Non-privacy policy related, place-holder or vague text} – 
The PrivPRISM framework processes all textual content from such pages and systematically evaluates them for meaningful privacy disclosures. Owing to its capacity to identify high-level data practice categories and fine-grained data items even in short text segments, PrivPRISM effectively detects these inadequate policies—typically reflected as very low PP compliance scores relative to DS declarations and static code evidence. Such cases constitute clear violations of Google Play’s developer policy and applicable regional privacy regulations, and are potentially actionable upon regulatory review.
Characteristic – \emph{Exhibits low PP compliance despite high DS compliance}}

\textcolor{black}{\textbf{02-	Boilerplate policies that are not-truly representative of app behaviour –}
This is one of the key artifacts highlighted in our study — the widespread reuse of generic PPs that are written to cover multiple apps or services. As a result, end-users are often overwhelmed by irrelevant or overly broad descriptions of data practices, making it difficult to discern what actually applies to a specific app. While boilerplate text does not necessarily imply non-compliance, such policies rarely align with the app’s real data handling behaviours and therefore undermine transparency and meaningful consent. Policy contradictions or non-representative policies are potentially actionable under regulatory and platform rules.
Characteristic – \emph{Exhibits low DS compliance despite high PP compliance}}

\subsection{Model Training}
\label{subsec:appendix_model_training}

During the designing and deployment of the PrivPRISM framework, we finetuned several language models locally. PrivBERT \cite{privbert} is $\sim100M$ parameter model available for research under a CC BY-NC-SA licensing. For fine-tuning and zero-shot prompting, we used meta-llama/Llama-3.1-8B and meta-llama/Llama-3.1-8B-Instruct\cite{grattafiori2024llama3herdmodels} available for research purposes under community licence agreement. Such model training and testing was conducted locally on two NVIDIA-RTX4090 computational nodes with parameter efficient fine tuning (PEFT) and low rank adaptation (LoRA) when necessary. For comparisons we also used GPT4o available via API access for which research usage is allowed under OpenAI Sharing and Publication Policy. \textcolor{black}{We observed the average inference time per privacy policy to be 4.85 minutes with a memory footprint of $\sim$41GB of VRAM. $\sim$75\% of this time was used for fine granular data practice decoding and the pipeline allows increasing the batchsize to reduce this percentage when provided with more capable hardware setups. For the encoder modules in our framework, the inference times were negligible. As an example, the self-supervised verifiers (batch size = 8; i.e., 8 inputs mapped to keywords) inferencing was <10s with a modest memory footprint of $\sim$2.4GB of GPU VRAM utilisation. }

\subsection{Acknowledgement}
This research was funded by Australian Research Council (ARC) Discovery Project DP220102520. We acknowledge the limited use of generative AI tools (ChatGPT) for text-refinement and grammatical checks during this manuscript drafting.

\end{document}